\documentclass[conference]{IEEEtran}
\IEEEoverridecommandlockouts
\PassOptionsToPackage{bookmarks={false}}{hyperref}
\usepackage{cite}
\usepackage[inline]{enumitem}
\usepackage{amsmath,amssymb,amsfonts}
\usepackage{subfig}
\usepackage{graphicx}
\usepackage{textcomp}
\usepackage{xcolor}
\usepackage{soul}
\usepackage{caption}
\usepackage{epstopdf}
\usepackage{array}
\usepackage{nicefrac}
\usepackage{siunitx}
\usepackage{adjustbox}

\newcommand{\tweak}{{\tt{Tweak}}} 

\newcommand{\widescan}{{\tt{WideScan}}}

\usepackage[colorinlistoftodos]{todonotes}
\makeatletter
\usepackage[utf8]{inputenc}
\newcommand{\inlineitem}[1][]{%
\ifnum\enit@type=\tw@
    {\descriptionlabel{#1}}
  \hspace{\labelsep}%
\else
  \ifnum\enit@type=\z@
       \refstepcounter{\@listctr}\fi
    \quad\@itemlabel\hspace{\labelsep}%
\fi}
\makeatother
\def\BibTeX{{\rm B\kern-.05em{\sc i\kern-.025em b}\kern-.08em
    T\kern-.1667em\lower.7ex\hbox{E}\kern-.125emX}}

\newcommand{\comment}[1]{ }

\usepackage{adjustbox}

\usepackage{lipsum}

\usepackage{amsmath}
\usepackage{float}
\usepackage{soul}
\usepackage{booktabs}
\usepackage{tabularx}
\usepackage{color}
\usepackage[algo2e]{algorithm2e}
\usepackage{lipsum,multicol}
\usepackage{array,multirow}
\usepackage{lettrine}
\usepackage{gensymb}
\usepackage{mathtools, nccmath}

\usepackage{algorithm}
\usepackage[font=footnotesize, labelfont={bf, color=accessblue}, margin=.75cm, labelsep = period]{caption}
\usepackage{algorithmic}
\usepackage{lscape}
\graphicspath{{./}{./Figure/}}
\usepackage{enumerate}
\usepackage{epstopdf}
\usepackage{float}
\usepackage{url}
\usepackage{booktabs}
\usepackage{cite}
\usepackage{array}

\newcommand{\myitemizebeginless}{\begin{list}{$\bullet$}
{
 \setlength{\leftmargin}{0.09cm}
 \setlength{\parsep}{0.0cm}
 \setlength{\itemsep}{0.05cm}
 \setlength{\topsep}{0.0cm}
}}
\newcommand{\myitemizeendless}{\end{list}}

\usepackage{tabularx}
\def\BibTeX{{\rm B\kern-.05em{\sc i\kern-.025em b}\kern-.08em T\kern-.1667em\lower.7ex\hbox{E}\kern-.125emX}}

\begin{document}


\title{Uncovering the Portability Limitation of Deep Learning-Based Wireless Device Fingerprints
\thanks{This article has also been accepted to 6G Summit, Abu Dhabi, UAE, Nov. 2022.
This work is supported in part by NSF/Intel Award No. 2003273.}
}

\author{\IEEEauthorblockN{Bechir Hamdaoui and Abdurrahman Elmaghbub}
\IEEEauthorblockA{{School of Electrical Engineering and Computer Science}, {Oregon State University}, Corvallis, Oregon, USA\\
\{hamdaoui,elmaghba\}@oregonstate.edu} }

\maketitle

\begin{abstract}
Recent device fingerprinting approaches rely on deep learning to extract device-specific features solely from raw RF signals to identify, classify and authenticate wireless devices. One widely known issue lies in the inability of these approaches to maintain good performances when the training data and testing data are collected under varying deployment domains. For example, when the learning model is trained on data collected from one receiver but tested on data collected from a different receiver, the performance degrades substantially compared to when both training and testing data are collected using the same receiver. The same also happens when considering other varying domains, like channel condition and protocol configuration. 
In this paper, we begin by explaining, through testbed experiments, the challenges these fingerprinting techniques face when it comes to domain portability. We will then present some ideas on how to go about addressing these challenges so as to make deep learning-based device fingerprinting more resilient to domain variability.

\end{abstract}

\begin{IEEEkeywords}
Device fingerprinting portability, deep learning, domain-adaptation, hardware impairments.
\end{IEEEkeywords}


\section{Introduction}
\label{sec:Introducation}
Recently, there has been considerable interest in adopting deep learning-enabled device fingerprinting in automated network authentication mechanisms for emerging large-scale wireless devices (e.g., 6G, IoT, vehicular, etc.)~\cite{hamdaoui2022deep,rajendran2022rf}.
%
In essence, device fingerprinting relies on deep learning techniques to extract device-specific features and signatures, solely from raw RF (radio frequency) signals, that can be used to uniquely identify devices. Deep learning is capable of automatically processing raw RF signals and exploiting random distortions in the received RF signals that are caused by the devices' hardware impairments to extract signatures that are unique to the devices~\cite{hamdaoui2020deep,sankhe2019oracle,elmaghbub2021lora}.
These hardware impairments occur during manufacturing of the transmitters' various hardware components, including the local oscillator, the power amplifier, the mixer, and others. More detailed explanation and analysis of various hardware impairments can be found in~\cite{elmaghbub2020widescan,elmaghbub2021lora}.

\section{Fingerprint Portability Challenges}
\label{sec:challenges}
Deep learning-based device fingerprinting approaches have indeed achieved good performance results when both training and testing data are collected under the same channel conditions and network deployment configurations. However, they perform poorly when the data used during training and that used during testing are collected under varied network deployment setting, where a deployment setting could refer to a channel condition, a hardware receiver, a protocol configuration, a physical location, etc. Each of these are often referred to as {\em domain}, and the problem being raised is referred to as {\em fingerprint portability} or as {\em domain-adaptation} in the machine learning community. We will next demonstrate how the limited portability of fingerprints can impact device fingerprinting accuracy, and we do so using testbed experiments.

\subsection{Testbed and Data Collection Setup}
\label{subsec:testbed}
\begin{figure}
\centering
\subfloat[USRP B210]{
\includegraphics[width=.45\columnwidth]{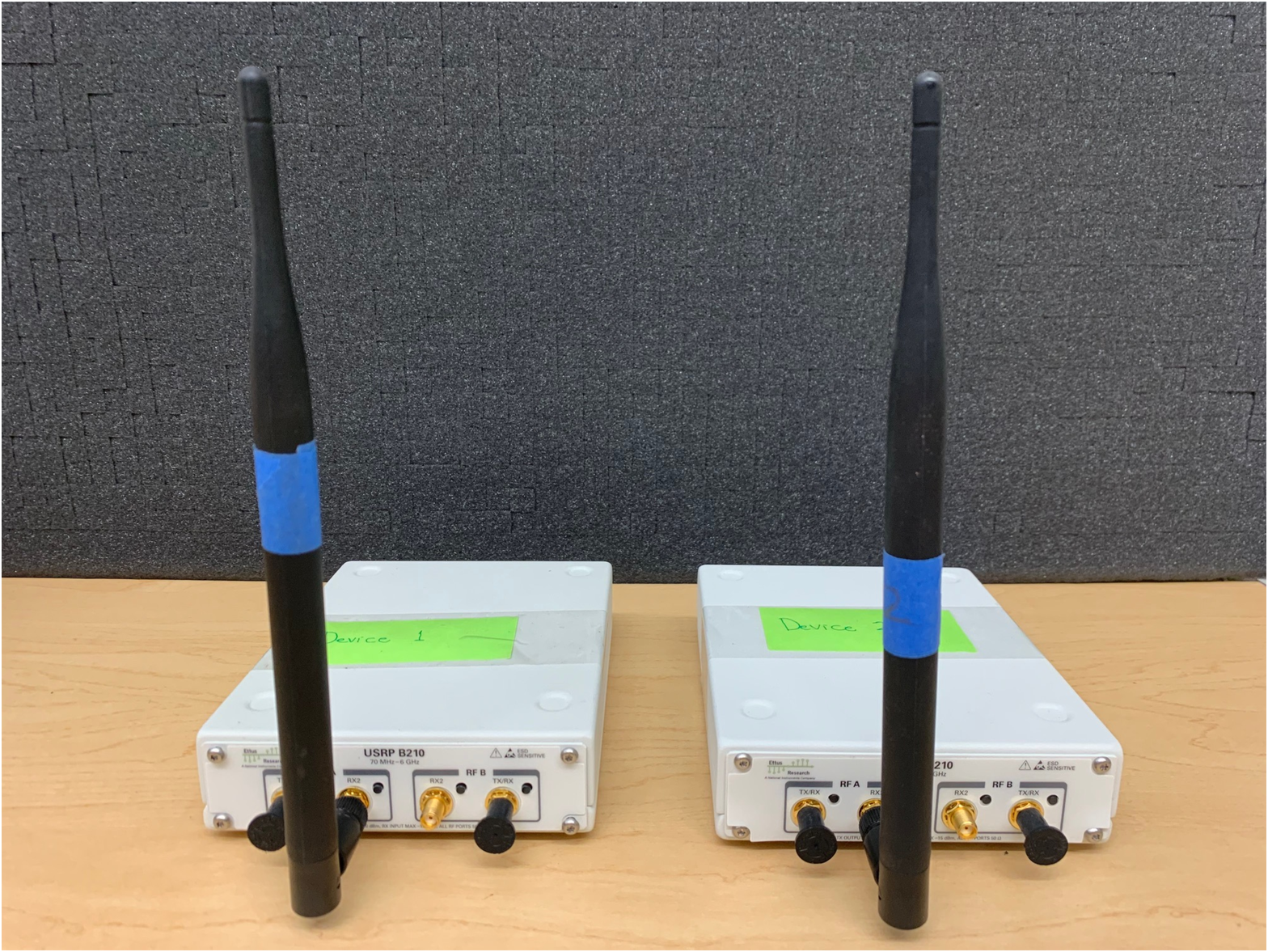}\label{subfig:usrp}}
\subfloat[PyCom devices]{
 \includegraphics[width=.52\columnwidth]{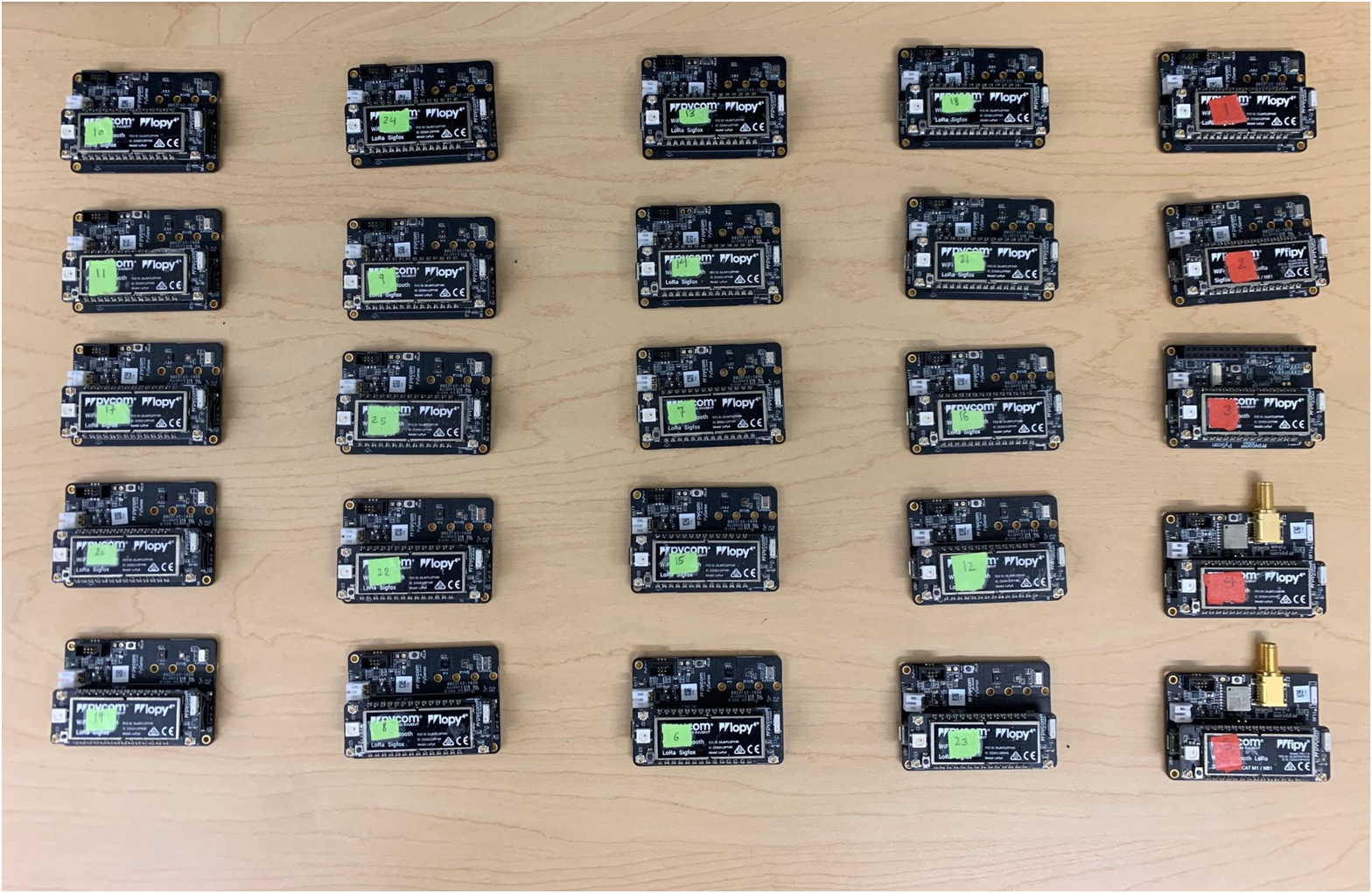}\label{subfig:tx}}
\caption{Experimental testbed~\cite{gaskin2022tweak}.}
\label{fig:testbed}
\end{figure}

To explain these challenges, we used our IoT fingerprinting testbed~\cite{elmaghbub2021comprehensive} to run several experiments under different varied domains, by training and testing the deep learning models on data collected on different days, using different receivers, and/or under different protocol configurations.
The IoT testbed, shown in Fig.~\ref{fig:testbed}, consists of 60 IoT/Pycom transmitters and 2 USRP B210 receivers, and is used to collect signal data under multiple different deployment settings, including indoors, outdoors and others. Each Pycom device was connected to a dedicated antenna and was configured to send LoRa packets at $125$KHz bandwidth in the $915$MHz frequency, with a  spreading factor set to $7$, a preamble set to $8$, and a transmit power level set to $20$dBm. For more detailed description of the testbed, experimental scenarios, and LoRa protocol specifications, see \cite{elmaghbub2021comprehensive,elmaghbub2021lora}.

\subsection{Results Illustrating Fingerprint Portability Challenges}
\begin{figure}
\centering
\subfloat[Channel portability]{
\includegraphics[width=0.97\columnwidth]{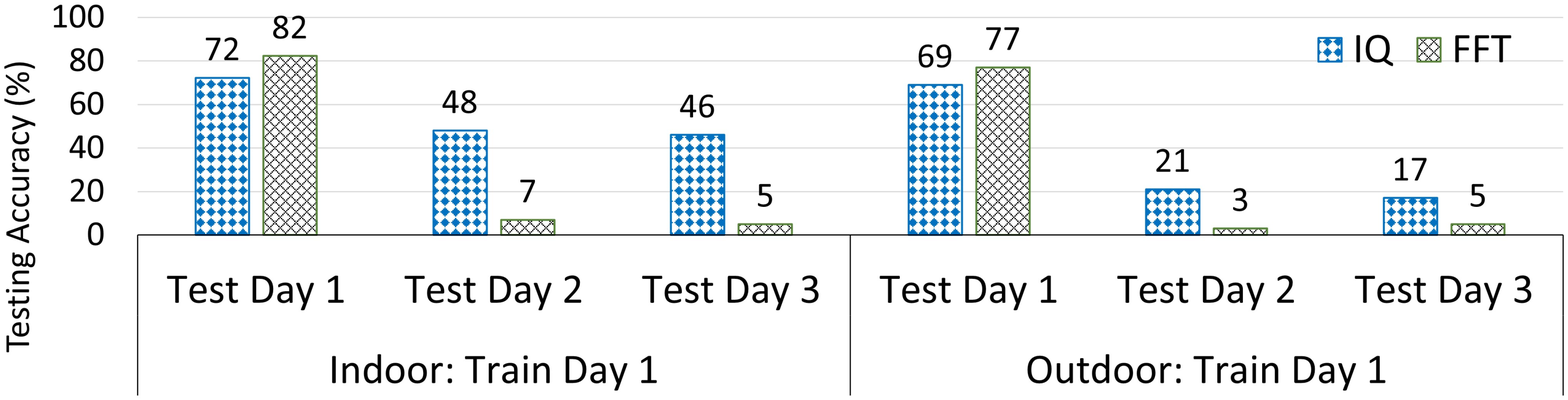}\label{subfig:time}}
   
\subfloat[Receiver hardware portability]{
\includegraphics[width=0.97\columnwidth]{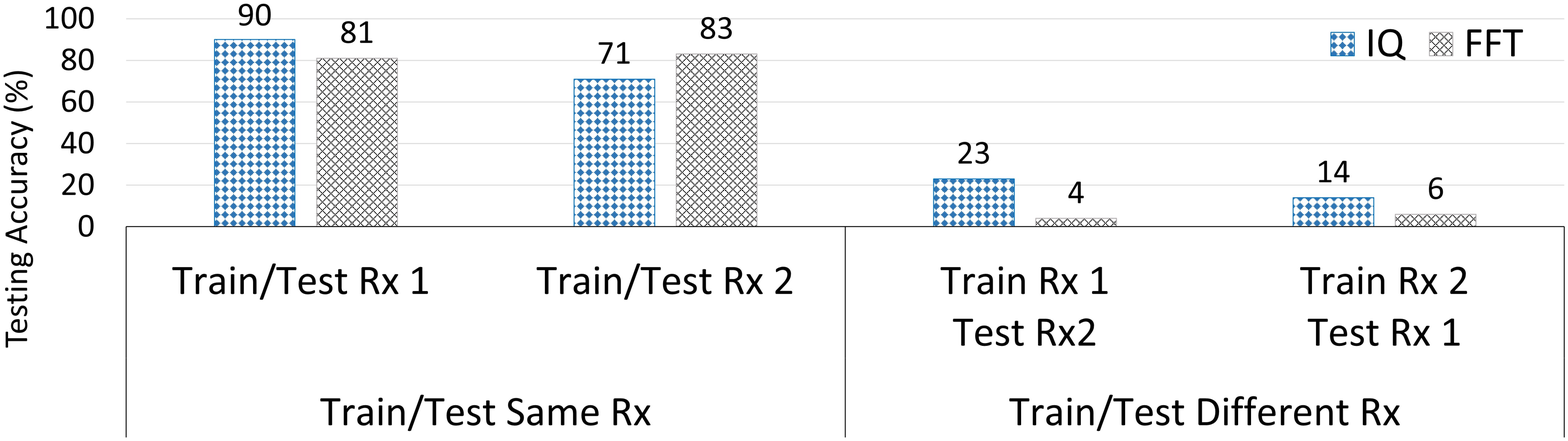}\label{subfig:rx}}
  
\subfloat[Protocol configuration portability]{
 \includegraphics[width=0.97\columnwidth]{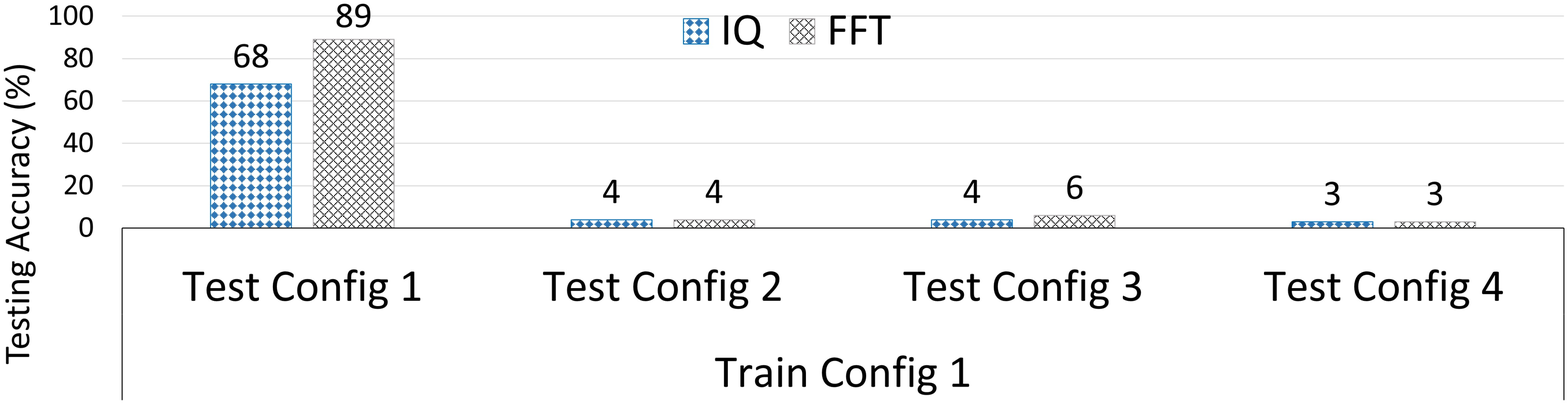}\label{subfig:config}}
\caption{Domain portability challenges when changing domain is (a) time, (b) receiver, or (c) configuration.}
\label{fig:challenges}
\end{figure}

Fig.~\ref{fig:challenges} demonstrates the impact of the poor fingerprint portability on the achieved accuracy of the learning models under three different varied domains: channel variability, receiver hardware variability, and protocol configuration variability. Specifically, Fig.~\ref{subfig:time} reveals the sensitivity of fingerprinting accuracy to channel variability, i.e., when training data and testing data are captured on different days. Fig.~\ref{subfig:rx} shows its sensitivity to receiver hardware variability, i.e., when training data and testing data are collected using different receivers. And Fig.~\ref{subfig:config} shows its sensitivity to transmitter protocol configuration variability, i.e., when training data is collected when the transmitting devices use one configuration, but the testing data is collected when these devices use a different protocol configuration. The figures clearly show the significant degradation in accuracy due to a change in the domain, whether the domain being channel condition, receiver hardware, or protocol configuration, and that  fingerprinting performs well only when the training and the testing domains (day, receiver, or protocol) match one another.


\section{Limitations of Prior Solution Attempts}
\label{sec:prior}
There have been some made efforts aimed at addressing the fingerprint portability challenges, ranging from adding artificial impairments to increase device separability~\cite{sankhe2019oracle}, to data augmentation to generalize the model to adapt to domain variation \cite{rajendran2022rf}, 
to modifying transmitted signals to compensate for channel condition variations~\cite{d2021can}, to using MIMO capability to overcome channel fading~\cite{basha2022leveraging}. The key issues with these proposed data augmentation approaches is that they require significant amounts of new labeled data, as well as re-training of the models with these new data. The other proposed approaches require modification at the transmitter side and are vulnerable to wireless channel conditions, as some require that receivers send their feedback to the transmitters in order for the transmitters to be able to modify their signals.

\section{Potential Solution Ideas}
\label{open-challenges}
We now highlight two techniques we proposed to (potentially) address the domain portability issues of device fingerprints. 

\begin{itemize}
\item \tweak~\cite{gaskin2022tweak}, a recently proposed technique that leverages metric learning to achieve accurate device identification in both closed-set and open-set scenarios. \tweak\ significantly reduces the training time as well the amount of data needed to do so. It achieves portability through a calibration process that is not computationally intensive and hence is employable in resource-constrained devices, and is accomplished with a limited amount of labeled training data from the target domain. 
Our experimental results using our IoT testbed described in Section~\ref{subsec:testbed} show that \tweak\ increases the resiliency of fingerprinting to changes in receiver hardware, wireless channel conditions, and the configuration of the transmitters' LoRa protocol.

\item \widescan~\cite{elmaghbub2020widescan,elmaghbub2021lora}, a technique we proposed to leverage signal distortions that occur due to the hardware impairments in the spectrum band surrounding the original bandwidth of the signal to uniquely identify devices. \widescan\ considers extracting features from both the in-band and the out-of-band spectrum information to increase device separability, thus increasing the resiliency of fingerprinting to domain variations. Our experimental results show that \widescan\ improves the accuracy of fingerprinting approaches.

\end{itemize}

\section{Concluding Remarks}
\label{sec:Conclusion}
This paper sheds some light on the fingerprints portability challenges, goes over some of the limitations of prior works, and covers a couple of proposed approaches. From our own research experience, it is clear that so much work still needs to be done to address the domain adaption and fingerprints portability problems in the wireless classification field.






\bibliographystyle{IEEEtran}

\bibliography{References}

\end{document}